%% file: ral.tex
\documentclass[letterpaper, 10 pt, conference]{ieeeconf}  

\IEEEoverridecommandlockouts                              

\usepackage{graphics} 
\usepackage{amsmath} 

\usepackage{amssymb}  
\usepackage{colortbl}	
\usepackage[table]{xcolor}
\usepackage[noend]{algpseudocode}
\usepackage{algorithm}
\usepackage[caption=true]{subfig}
\usepackage[pdftex]{graphicx}
\graphicspath{{./fig/biped/}, {./fig/}}
\DeclareGraphicsExtensions{.png,.pdf}

\newcommand{\changed}[2]{#1} 	

\input{./math_commands.tex}

\title{\LARGE \bf
Exponential Integration for Efficient and Accurate Multi-Body Simulation with Stiff Viscoelastic Contacts
}

\author{Bilal Hammoud$^{1}$, Luca Olivieri$^{2}$, Ludovic Righetti$^{1}$, Justin Carpentier$^{3}$, Andrea Del Prete$^{2}$
\thanks{$^{1}$Tandon School of Engineering, New York University, New York, USA. 
        {\tt\small bilal.hammoud@nyu.edu, ludovic.righetti@nyu.edu}}%
\thanks{$^{2}$Department of Industrial Engineering, University of Trento, Italy
        {\tt\small luca.olivieri-1@unitn.it, andrea.delprete@unitn.it}}%
\thanks{$^{3}$INRIA, Paris, France
        {\tt\small justin.carpentier@inria.fr}}%
}

\begin{document}

\maketitle
\thispagestyle{empty}
\pagestyle{empty}

\begin{abstract}
The simulation of multi-body systems with frictional contacts is a fundamental tool for many fields, such as robotics, computer graphics, and mechanics. 
Hard frictional contacts are particularly troublesome to simulate because they make the differential equations stiff, calling for computationally demanding implicit integration schemes.
We suggest to tackle this issue by using exponential integrators, a long-standing class of integration schemes (first introduced in the 60's) that in recent years has enjoyed a resurgence of interest.
We show that this scheme can be easily applied to multi-body systems subject to stiff viscoelastic contacts, producing accurate results at lower computational cost than \changed{classic explicit or implicit schemes}.
In our tests with quadruped and biped robots, our method demonstrated stable behaviors with large time steps (10 ms) and stiff contacts ($10^5$ N/m).
Its excellent properties, especially for fast and coarse simulations, make it a valuable candidate for many applications in robotics, such as simulation, Model Predictive Control, Reinforcement Learning, and controller design.
\end{abstract}


\input{sections/intro.tex}
\input{sections/method.tex}
\input{sections/results.tex}
\input{sections/conclusions.tex}

\addtolength{\textheight}{-0cm}   

%

\bibliographystyle{IEEEtran}
\bibliography{references_clean}

\end{document}

%% file: math_commands.tex
\usepackage{bm} 

\newcommand{\pder}[2]{\frac{\partial#1}{\partial#2}}		
\newcommand{\mat}[1]{\ensuremath{\begin{bmatrix}#1\end{bmatrix}}}	

\newcommand{\dt}[0]{\ensuremath{\delta t}}					
\newcommand{\Rv}[1]{\ensuremath{\mathbb{R}^{#1}}}				
\newcommand{\Rm}[2]{\ensuremath{\mathbb{R}^{#1\times #2}}}		
\newcommand{\Expect}{{\rm I\kern-.3em E}}				

\newcommand{\Cone}{\ensuremath{\mathcal{K}_{\mu}}}

\newenvironment{eqs*}{\begin{equation}\aligned}{\endaligned\end{equation}}
\newenvironment{eqs}[1]{\equation\label{#1}\aligned}{\endaligned\endequation}

%% file: sections/intro.tex
\section{Introduction}
The interest of the robotics community for fast and reliable methods to simulate multi-body systems subject to frictional contacts has been constantly growing in the last two decades~\cite{Anitescu2004,Todorov2010,Todorov2014,Hwangbo2018,Drumwright2019}.
This is reasonable considering that simulation is at the core of many robotics applications, such as the development and testing of novel controllers before deployment on hardware.
Moreover, many advanced control and planning techniques, such as Model Predictive Control~\cite{Tassa2012} (MPC) and Optimal Control~\cite{Stryk1993}, rely on the ability to predict the future behavior of the system.
Finally, the current bottleneck of many learning algorithms~\cite{Mansard2018, Viereck2018} is their need for huge amounts of data, which therefore can greatly benefit from fast and accurate simulation methods.

The simulation of articulated rigid multi-body systems without contacts is a solved problem~\cite{featherstone2014rigid}.
The same is not the case for systems with stiff contacts, which can be treated in two ways, each leading to a hard (but different) numerical problem.
The first approach, which more closely follows the physical phenomenon of contacts, consists in expressing contact forces as a function of the penetration between bodies.
Often, linear spring dampers have been used in this context~\cite{Drumwright2019,Yamane2006}.
This leads to stiff differential equations that are simple to evaluate, but difficult to integrate because of their \emph{numerical stiffness}~\cite{Ascher1998}.
The second approach tries to circumvent these numerical challenges by assuming contacts to be \emph{infinitely} rigid.
This approach effectively gets rid of the numerical stiffness, but in exchange for \emph{non-smoothness}.
One method that has been particularly successful for dealing with the resulting non-smooth equations is velocity-impulse time-stepping~\cite{Anitescu,Anitescu2004,Hwangbo2018,peiret2019}.
This has become the standard for robot simulation~\cite{Erez2015}, demonstrating stable behaviors with large time steps (several ms)---even though it must solve a numerically hard Linear Complementarity Problem (LCP).
Several authors have tried to improve this approach by getting rid of the strict complementarity constraints~\cite{Todorov2010,Drumwright2010,Todorov2014}, which are the source of the numerical challenge.
However, none of these approaches is currently widely accepted in the robotics community, mainly because of the unclear effects of the introduced numerical regularization/relaxations on the physics (i.e., relaxations can be interpreted as implicit spring/dampers but are not an explicit part of the modeling).

\begin{figure}[!tbp]
   \centering
   \includegraphics[width=0.7\linewidth]{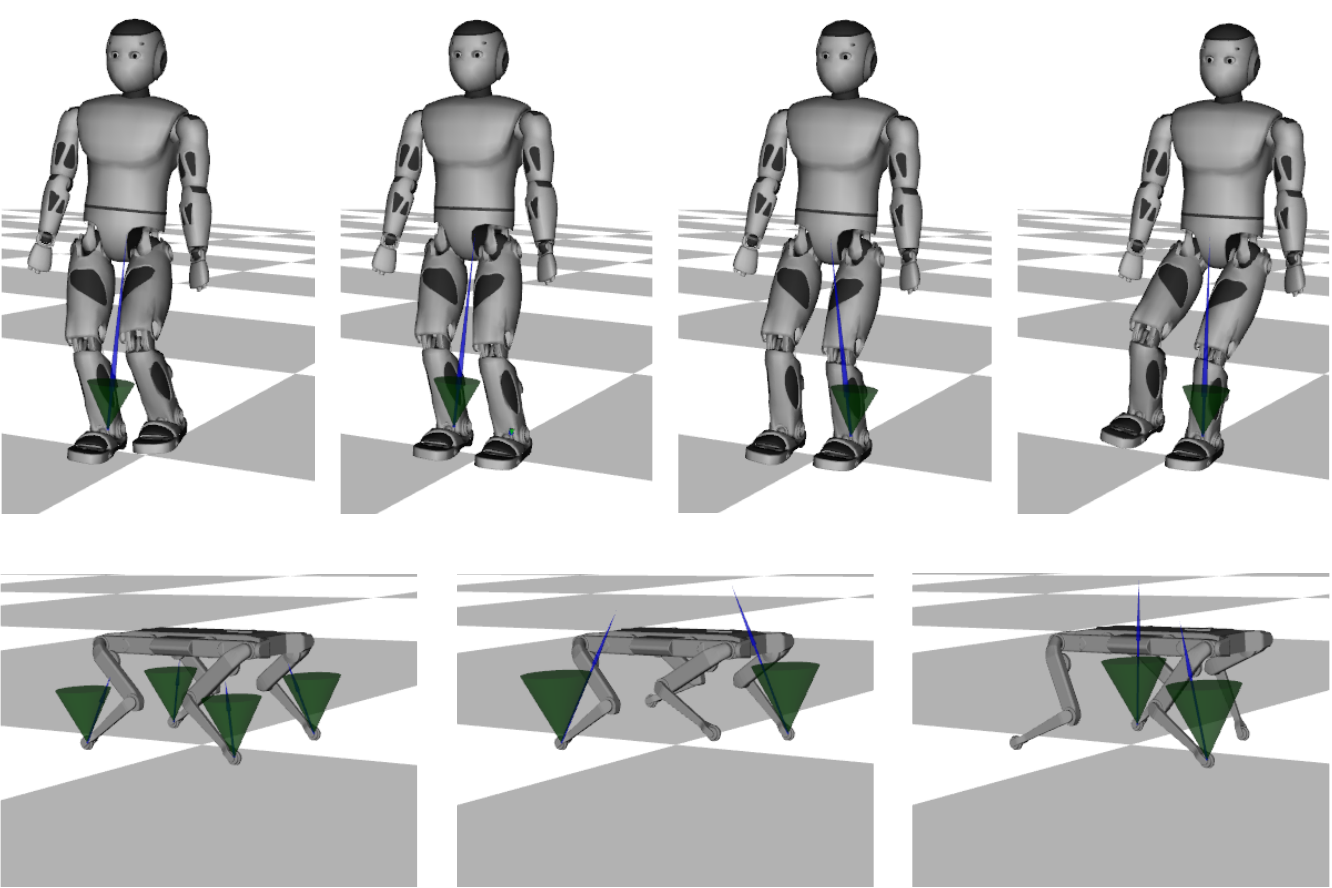} 
   \caption{Snapshots from our simulation tests with a biped and a quadruped robot.}
   \label{fig:snapshots}
\end{figure}

The approach we advocate for in this article is based on a well-known soft contact model: the linear spring damper.
Instead of using explicit integration schemes, which require small time steps, or implicit schemes, which require solving nonlinear systems of equations and introduce artificial viscosity leading to nonphysical behaviors, we use \emph{Exponential Integrators} (EI)~\cite{Loffeld2013, chen2020}.
EI are a long-standing class of integration schemes~\cite{Certaine1960} that are particularly suited for stiff differential equations.
EI were initially considered unpractical because of the computational challenges related to the matrix exponential~\cite{Moler2003}.
However, novel numerical methods to compute the matrix exponential~\cite{Higham2005,AL-MOHY2011,Sastre2019} have recently unlocked the potential of EI.
This has already been used in computer graphics for simulating deformable objects, modeled as systems of particles~\cite{Michels2014, Chen2018, May2018}.
This model is particularly suited for EI because the stiff part of the dynamics is linear, which however is not the case for articulated systems in contact with the environment.

Our main contribution is a simulation algorithm that exploits EI to simulate articulated robots in contact with a stiff visco-elastic environment. 
Particularly, this paper addresses the following questions. 
(1) Does the proposed simulation scheme provide an improvement in terms of speed vs accuracy when compared to classic explicit and implicit methods?
(2) Is it possible to develop a simulation scheme that is less sensitive to the choice of contact stiffness and damping? 
(3) Is it possible to get stable simulations with increased time step (integration interval)? 
The last question is of particular importance for using MPC and reinforcement learning.     

In order to address all the above questions in an efficient way, we apply EI only to the contact dynamics (which is stiff) while using an explicit Euler scheme for the remaining terms (which are not stiff).
Our simulation results on quadruped and biped robots (see Fig.~\ref{fig:snapshots}) show the superior performance of our method compared to \changed{standard integration schemes}\~ in terms of accuracy, speed and stability.
To our knowledge, this class of integrators has never been used before by the robotics community.
The article is organized as follows. 
Section~\ref{sec:background} introduces the problem of multi-body simulation and the basic theory of EI.
Section~\ref{sec:method} explains how EI can be used for multi-body simulation with bilateral contacts.
This method is then extended to frictional contacts in Section~\ref{sec:friction}.
Section~\ref{sec:computation} discusses the implementation details of the algorithm.
Section~\ref{sec:results} presents the results and Section~\ref{sec:conclusions} concludes the article.

%% file: sections/method.tex
\section{Background}
\label{sec:background}
\subsection{Multi-Body Dynamics and Soft Contact Model}
We want to simulate a multi-body mechanical system with the following dynamics~\cite{featherstone2014rigid}:
\begin{eqs}{eq:robot_dyn}
M(q) \dot{v} = u(q,v) + J(q)^\top \lambda,
\end{eqs}
where $q$ is the robot joint configuration, $v$ is the robot joint velocity, $M$ is the joint space mass matrix, $u$ contains gravity, nonlinear effects and actuator forces, $J$ is the contact point Jacobian, and $\lambda$ contains the stacked 3D contact forces.
We assume a linear spring-damper contact model, which means that the contact forces $\lambda$ are proportional to the inter-penetration of contacting bodies:
\begin{eqs}{}
\lambda = - K \, (\underbrace{p(q) - p_0}_{\Delta p}) - B \, (\underbrace{\dot p(q, v) - \dot{p}_0}_{\Delta \dot p}),
\end{eqs}
where $p$ and $\dot{p}$ contain the stacked 3D contact point positions and velocities, $p_0$ and $\dot{p}_0$ contain the stacked 3D anchor point positions and velocities, and $K$ and $B$ are the diagonal stiffness and damping matrices, respectively.
The anchor point $p_0$ is a \emph{virtual} point to which the \emph{virtual} spring and damper are attached.
It is typically set to the contact point location when contact is first detected, and as long as contacts are sticking $\dot{p}_0 = 0$.
However, when slipping occurs then $\dot{p}_0 \neq 0$.
A limitation of the ``anchor point'' model is that to generate tangential forces, some lateral motion of the contact point is always necessary. Consequently, pure static friction cannot be modeled with this approach, but it can be well approximated by using large lateral stiffnesses.
Dependencies on $q$ and $v$ are dropped in the following to ease notation.

\subsection{Explicit Integration Schemes}
The classic approach to integrate this dynamical system starts by writing it in standard form.
Defining the state as $x_q \triangleq (q, v)$, its dynamics is:
\begin{eqs}{eq:nlds}
\underbrace{\frac{d}{dt} \mat{q \\ v}}_{\dot x_q} = \underbrace{\mat{v \\ M^{-1} (u + J^\top \lambda)}}_{f(x_q, u)}
\end{eqs}
We can integrate~\eqref{eq:nlds} with any numerical integration scheme, such as a high-order Runge-Kutta scheme, or even a simple explicit Euler~\cite{Ascher1998} (very common in robotics):
\begin{eqs}{}
x_q^+ = x_q + \dt \, f(x_q, u)
\end{eqs}
where $x_q^+$ represents the next value of the state and $\dt$ is the integration time step.
The problem with this approach is that for large values of $K$ and $B$ the differential equations~\eqref{eq:nlds} are \emph{stiff}~\cite{Ascher1998}. 
This means that they require very small integration steps for numerical stability.
This is the main reason why soft contact models have been mostly abandoned in the last decade, in favor of complementarity-based models (and their relaxations) and time-stepping integration~\cite{Todorov2010,Todorov,Hwangbo2018}.

\subsection{Exponential Integrators (EI)}
EI~\cite{Chen2018,May2018,Michels2014} are integration schemes particularly suited for stiff dynamical systems for which the \emph{stiffness} comes from a linear part of their dynamics:
\begin{eqs}{eq:exp_int_dyn}
\dot x = \underbrace{f(x)}_{\text{nonstiff nonlinear function}} + \underbrace{A x}_{\text{stiff linear function}}
\end{eqs}
In this case, using an explicit integration scheme would result in the problems mentioned above.
Instead, EI exploit the linearity of the stiff part of the dynamics, which can be solved \emph{analytically} using the matrix exponential, thanks to the well-known solution of linear dynamical systems:
\begin{eqs}{eq:lds_sol}
\dot x(t) &= A x(t) + b \\ 
x(t) &=  e^{t A} x(0) + \int_0^t e^{\tau A}  \, \text{d}\tau \, b
\end{eqs}
First-order EI apply the solution~\eqref{eq:lds_sol} to the nonlinear system~\eqref{eq:exp_int_dyn}, by interpreting $f()$ as $b$ and assuming it remains constant during the integration step:
\begin{eqs}{eq:nlds_sol}
x(t) = e^{t A} x(0) + \int_0^t e^{\tau A}  \text{d}\tau \, f(x(0))
\end{eqs}
Since the stiff part of the equations is integrated via the matrix exponential, large integration steps can be taken. 

\section{Bilateral Contacts}
\label{sec:method}
Our approach consists in using EI to simulate the system~\eqref{eq:nlds}.
The standard approach to apply EI to arbitrary dynamics is to use a 1-st order Taylor expansion:
\begin{eqs}{}
\dot{x}_q(t_0+t) \approx \dot{x}_q(t_0) + \pder{f}{x_q} \, (x_q(t_0+t)-x_q(t_0))
\end{eqs}
However, this would require two demanding computations: the dynamics Jacobian, and a matrix exponential with the size of $x_q$.
In the following we present instead an approach that i) does not require the dynamics Jacobian, and ii) only computes a matrix exponential with twice the size of $\lambda$, which is typically smaller than $x_q$ in legged locomotion. 

To get a differential equation with the form~\eqref{eq:exp_int_dyn} we start by projecting~\eqref{eq:robot_dyn} in contact space pre-multiplying both sides by $J M^{-1}$:
\begin{eqs}{}
J \dot{v} - \underbrace{J M^{-1} J^\top}_{\Upsilon} \lambda = J \underbrace{M^{-1} u}_{\dot{\bar{v}}}
\end{eqs}
Then we use the relationship $\ddot p = J \dot v + \dot J v$ to express the contact point accelerations as functions of the robot accelerations:
\begin{eqs}{}
\ddot p - \Upsilon \lambda = J \dot{\bar{v}} + \dot J v \\
\ddot p + \Upsilon K \Delta p + \Upsilon B \Delta \dot p = \underbrace{J \dot{\bar{v}} + \dot J v}_{\ddot{\bar{p}}} \\
\end{eqs}
Since for bilateral contacts $\dot{p}_0$ is always null, we have that $\ddot{p}_0=0$ and thus we can write the contact point dynamics as:
\begin{eqs}{eq:contact_dyn}
\frac{d}{dt} \mat{\Delta p\\ \Delta \dot p} = \underbrace{\mat{0 & I \\ -\Upsilon K & -\Upsilon B}}_A \underbrace{\mat{\Delta p \\ \Delta \dot p}}_x + \underbrace{\mat{0 \\ \ddot{\bar{p}}}}_b
\end{eqs}
This dynamical system does not have the same form as~\eqref{eq:exp_int_dyn} because $\Upsilon$ (and thus $A$) depends on $q$.
However, $\Upsilon$ is typically a well-conditioned function, meaning that it changes little for small variations of $q$. 
The same holds for $\ddot{\bar{p}}$ (and thus $b$), which is why multi-body systems without contacts can typically be integrated with large time steps ($\approx$5 ms).
This means that we can approximate $A$ and $b$ as constants during the integration step, and therefore treat~\eqref{eq:contact_dyn} as linear.
We can now express the contact forces as:
\begin{eqs}{eq:contact_forces}
\lambda(t) &= \underbrace{\mat{-K & -B}}_D x(t) = D e^{t A} x(0) + D \int_0^t e^{\tau A}  \text{d}\tau \, b
\end{eqs}

Substituting~\eqref{eq:contact_forces} in~\eqref{eq:nlds} we can compute the robot accelerations:
\begin{eqs}{}
\dot v(t) &= M^{-1} (u + J^\top \lambda(t)) = \dot{\bar{v}} + M^{-1} J^\top D x(t),
\end{eqs}
where we consider all terms constant during the integration step, except for $x(t)$.
Now we can integrate to get the new velocities $v^+$:
\begin{eqs}{eq:vel_int}
v^+ &= v + \int_0^{\dt} \dot v(\tau) \, \text{d}\tau = \\
&= v + \dt \, \dot{\bar{v}} + M^{-1} J^\top D \int_0^{\dt} x(\tau)  \, \text{d}\tau
\end{eqs}
Finally we integrate twice to get the new configuration $q^+$:
\begin{eqs}{eq:pos_int}
q^+ &= q + \int_0^{\dt} v(\tau) \text{d}\tau = \\
&= q + \dt \, v + \frac{{\dt}^2}{2} \dot{\bar{v}} + M^{-1} J^\top D \int_0^{\dt} \int_0^{\tau} x(\tau_1)  \, \text{d}\tau_1 \, \text{d}\tau
\end{eqs}

\subsection{Integration of Matrix Exponentials}
Eq.~\eqref{eq:vel_int} and~\eqref{eq:pos_int} are straightforward to compute, except for their last terms, which are:
\begin{eqs}{eq:x_int_def}
x_{int}(t) &\triangleq \int_0^{t} x(\tau)  \, \text{d}\tau \\
x_{int2}(t) &\triangleq \int_0^{t} \int_0^{\tau} x(\tau_1)  \, \text{d}\tau_1 \, \text{d}\tau,
\end{eqs}
where
\begin{eqs}{eq:lds_sol2}
x(t) &=  e^{t A} x(0) + \int_0^t e^{\tau A}  \, \text{d}\tau \, b
\end{eqs}
When $A$ is invertible we can 
express the integral of $e^{tA}$ as an algebraic function of $e^{tA}$:
\begin{eqs}{eq:int_mat_exp_rel2}
\int_0^t e^{\tau A} \text{d}\tau = A^{-1} (e^{tA} - I)
\end{eqs}
However, $A$ is not invertible if the contact Jacobian $J$ is not full-row rank.
Luckily, the computation of integrals involving matrix exponentials has been thoroughly investigated~\cite{VanLoan1978,Carbonell2008}.
In Section~\ref{sec:computation} we show how to compute these integrals indirectly, by simply computing the matrix exponentials of an augmented system.

\subsection{Extension to non-Euclidian spaces}
When $q$  does not belong to an Euclidian space (as in the case of legged robots, where $q$ includes the orientation of the base link) the integration of $q$ is slightly more complicated (while the integration of $v$ remains unchanged).
Given the following definition:
\begin{eqs}{}
v_{mean} &\triangleq v + \frac{\dt}{2} \dot{\bar{v}} + \frac{1}{\dt} M^{-1} J^\top D x_{int2}(\dt)
\end{eqs}
The integration step of $q$ is computed as:
\begin{eqs}{}
q^+ &= \text{integrate}(q, \dt \, v_{mean}),
\end{eqs}
where the function \emph{integrate($\cdot$)} performs integration in the non-Euclidian space of $q$.

\section{Frictional Contacts}
\label{sec:friction}
So far we have assumed that contact forces were bilateral.
However, we typically want to simulate unilateral contacts, where forces oppose penetration but do not oppose detachment of bodies.
Assuming that the contact forces are expressed in a local reference frame with the z direction aligned with the contact normal, unilateral forces must satisfy:
\begin{eqs}{eq:unilateral_constr}
f_i^z \ge 0 \qquad \forall i
\end{eqs}
Moreover, tangential forces are typically limited as well. 
Assuming a Coulomb friction model we have:
\begin{eqs}{eq:friction_constr}
\sqrt{(f_i^x)^2 + (f_i^y)^2} \le \mu f_i^z \qquad \forall i,
\end{eqs}
where $\mu \in \mathbb{R}^{+}$ is the coefficient of friction\footnote{We assume that static and dynamic coefficients of friction are equal.}.
We can represent the constraints~\eqref{eq:unilateral_constr} and~\eqref{eq:friction_constr} as $\lambda \in \Cone$, with \Cone\ being a second-order cone.

\subsection{Force Projection}
To account for these constraints, when the value of $\lambda(t)$ computed by~\eqref{eq:contact_forces} is outside \Cone, we should project it on the boundaries of \Cone.
However, we do not know how to check this constraint in continuous time.
In the same spirit of time-stepping simulators~\cite{Anitescu2004}, we suggest to check friction constraints on the average value of $\lambda(t)$ during the integration step, which is:
\begin{eqs}{}
\bar{\lambda} \triangleq \frac{1}{\dt} \int_0^{\dt} \lambda(\tau) \, \text{d}\tau = \frac{1}{\dt} D x_{int}(\dt)
\end{eqs}
If $\bar{\lambda} \notin \Cone$, then we compute its projection on the boundaries of the friction cone $\lambda_{pr} = \text{proj}_{\Cone}(\bar{\lambda})$ and we use it to compute the next state:
\begin{eqs}{eq:state_int_friction}
\dot{v}_{pr} \triangleq M^{-1} (u + J^\top \lambda_{pr}) \\
v^+ = v + \dt \, \dot{v}_{pr}, \qquad q^+ = q + \dt \, v + \frac{\dt^2}{2} \dot{v}_{pr}
\end{eqs}
Note that in case $\bar{\lambda} \in \Cone$, then $\lambda_{pr} = \bar{\lambda}$ and the velocity update in~\eqref{eq:state_int_friction} is equivalent to~\eqref{eq:vel_int}. However, the position update in~\eqref{eq:state_int_friction} approximates the double integral of $\lambda(t)$ assuming a constant force ($\lambda_{pr}$), and so it is not equivalent to~\eqref{eq:pos_int} in general.
In order to exploit also the double integral of $x(t)$, we can check the friction cone constraints on the average of the average $\lambda(t)$, computed as:
\begin{eqs}{}
\bar{\bar{\lambda}} \triangleq \frac{2}{\dt^2} \int_0^{\dt} \int_0^{\tau} \lambda(\tau_1) \, \text{d}\tau_1 \text{d}\tau = \frac{2}{\dt^2} D x_{int2}(\dt)
\end{eqs}
If $\bar{\bar{\lambda}} \notin \Cone$, then we project it on the boundaries of the friction cone $\lambda_{pr2} = \text{proj}_{\Cone}(\bar{\bar{\lambda}})$ and we use it to compute the next position:
\begin{eqs}{eq:state_int_friction2}
\dot{v}_{pr2} &\triangleq M^{-1} (u + J^\top \lambda_{pr2}) \\
q^+ &= q + \dt \, v + \frac{\dt^2}{2} \dot{v}_{pr2}
\end{eqs}
Using \eqref{eq:state_int_friction2} for the position update and \eqref{eq:state_int_friction} for the velocity update, both updates are equivalent to the original ones in case of no slippage.

\subsection{Anchor point update}
When slippage occurs, the tangent anchor point state $(p_0^t, \dot{p}^t_0)$ (where the index $t$ indicates the tangent directions) changes, which has two main implications.
First, the assumption $\ddot{p}_0=0$ that we took to write the contact point dynamics as \eqref{eq:contact_dyn} is no longer valid.
This means that, during slippage, \eqref{eq:contact_dyn} is an approximation of the contact point dynamics, based on a ``business as usual'' assumption (i.e., that the anchor point $p_0$ continues slipping at constant velocity).
Second, the anchor point state should be updated so that the contact forces at the end of the time step are inside the friction cones.
When a contact is slipping, the tangent anchor point velocity converges to $\dot{p}^t$.
We show it now for the case of a 2D contact, but a similar reasoning can be applied to the 3D case.
While a contact is slipping, the tangential force $\lambda^t$ remains on the boundary of the friction cone, so we have:
\begin{eqs}{}
\dot{\lambda}^t &= \mu \dot{\lambda}^n \\
K^t (\dot{p}_0^t - \dot{p}^t) + B^t (\ddot{p}_0^t - \ddot{p}^t) &= \mu \dot{\lambda}^n \\
(\ddot{p}_0^t - \ddot{p}^t) &= - (B^t)^{-1} K^t (\dot{p}_0^t - \dot{p}^t) + \mu (B^t)^{-1} \dot{\lambda}^n
\end{eqs}
The last equation shows that, if $\dot{\lambda}^n = 0$, we have an exponential convergence to zero of $(\dot{p}_0^t - \dot{p}^t)$, with rate $(B^t)^{-1} K^t$.
Since typically $(B^t)^{-1} K^t$ is large, whereas $\mu (B^t)^{-1} \dot{\lambda}^n$ is small, we can expect this convergence to be fast.
For instance, if $(B^t)^{-1} K^t = 10^3$ and $\dot{\lambda}^n = 0$, then after 3 ms $(\dot{p}_0^t - \dot{p}^t)$ will be $5\%$ of its initial value.
Given this fast convergence, we neglect the transient and as soon as slippage starts we set $\dot{p}_0^t := \dot{p}^t$.
Then, we compute $p_0^t$ so that the contact force is on the boundary of the friction cone:
\begin{eqs}{}
\lambda &:= \text{proj}_{\Cone}(\lambda) \\
p_0^t &:= p^t + (K^t)^{-1} \lambda^t
\end{eqs}

\section{Computational Aspects}
\label{sec:computation}
The computational bottleneck of the presented approach is the computation of $x_{int}$ and $x_{int2}$ defined in~\eqref{eq:x_int_def}.
This section shows how to compute these quantities with a matrix exponential, and how this computation can be sped up.

\subsection{Computing $x_{int}$ and $x_{int2}$}
Using the results presented in~\cite{Carbonell2008} we can compute $x_{int}$ and $x_{int2}$ as:
\begin{eqs}{eq:x_int_computation}
\mat{x_{int}(t) & x_{int2}(t)} = \mat{I_n & 0_{n\times 3}} e^{t \bar A} \mat{0_{(n+1) \times 2} \\ I_2}
\end{eqs}
where $n$ is the size of $A$, and $\bar{A}\in \Rm{(n+3)}{(n+3)}$ is an augmented matrix:
\begin{eqs}{}
\bar A \triangleq 
\mat{A & b & x(0) & 0 \\ 
        0 & 0 & 1     & 0 \\
        0 & 0 & 0     & 1 \\
        0 & 0 & 0     & 0}
\end{eqs}

\subsection{Computing the Matrix Exponential}
Using~\eqref{eq:x_int_computation} we have transformed the problem of computing~\eqref{eq:x_int_def} into a matrix exponential evaluation.
Computing the matrix exponential is a challenging but well-understood numerical problem~\cite{Higham2005,Al-Mohy2010,AL-MOHY2011,Sastre2019}.
We have used as starting point the scaling\&squaring method, as revisited by Higham~\cite{Higham2005}, a widely used method for computing the exponential of small-medium size dense matrices.
The method scales the matrix by a power of 2 to reduce the norm to order 1, computes a Pad\'e approximant to the matrix exponential, and then repeatedly squares to undo the effect of the scaling.
A Pad\'e approximant of a function is its ``best'' approximation achievable by a ratio of two polynomials $D_j(\cdot)$, $N_j(\cdot)$ of order $j$:
\begin{eqs}{}
e^{A} \approx D_j(A)^{-1} \, N_j(A)
\end{eqs}
These approximants are only accurate around zero, so they cannot be used directly if $\|A\|$ is large.
When that is the case, the scaling\&squaring method is used to reduce $\|A\|$ by exploiting this property of the exponential:
\begin{eqs}{}
e^A = (e^{A/(2^s)})^{2^s}
\end{eqs}
The integer scaling parameter $s$ is chosen so that $\|e^{A/(2^s)}\|$ is sufficiently small.

\subsection{Boosting the Matrix Exponential Computation} \label{ssec:boost_expm}
Our problem has two features that we can exploit to speed up computation:
\begin{enumerate}
\item We do not need double machine precision, i.e. \mbox{$\approx10^{-16}$}, (which is the target of the algorithm of~\cite{Higham2005}) because we are typically fine with much larger numerical integration errors, e.g. $\approx10^{-4}$.
\item We do not need the whole matrix exponential, but only its product with a 2-column matrix, as shown in~\eqref{eq:x_int_computation}.
\end{enumerate}
The first point is easily exploitable. 
The choice of the scaling parameter $s$ and the polynomial order $j$ is usually optimized to achieve double machine precision with the minimum amount of matrix-matrix multiplications. 
We have empirically found that for our tests we can set $s=0$ and use a relatively low order $j \in [1, 2, 3, 5, 7]$, corresponding to $[0, 1, 2, 3, 4]$ matrix-matrix multiplications, respectively.
Which polynomial order is optimal depends on the specific test, and is discussed in the next section.

Regarding the second point, given a matrix $V$, we can directly compute the product $e^{A} V$ by performing operations in the following order:
\begin{eqs}{}
V_1 &:= N_j(A) \,V \\
e^A V &:= D_j(A)^{-1} \, V_1
\end{eqs}
This is faster than computing $e^A$ and then multiplying it times $V$ because we have to solve the linear system with a much smaller right-hand-side ($V_1$ rather than $N_j$).

Finally, we have also observed that the preprocessing step suggested in~\cite{Higham2005}, which uses \emph{matrix balancing}, is extremely effective at reducing $\|A\|$ in our tests.
This is crucial to achieve accurate results with low polynomial orders, therefore speeding up computations.
Further details can be found in our open-source online repository\footnote{https://github.com/andreadelprete/consim}.

%% file: sections/results.tex
\section{Results}
\label{sec:results}
We assess the performance of our simulation algorithm (\emph{Expo}) comparing it to Implicit Euler (\emph{Eul-imp}), Runge-Kutta 4 (\emph{RK4}) and explicit Euler (\emph{Eul-exp}).
Our implementation of \emph{Eul-imp} is described in the Appendix. 
Our implementation of \emph{RK4} is standard, whereas \emph{Eul-exp} was implemented as follows:
\begin{eqs}{}
v^+ = v + \dt \, \dot{v}, \qquad
q^+ = \text{integrate}(q, \dt \, v + \frac{\dt^2}{2} \, \dot{v})
\end{eqs}
Our results try to answer to the following questions:
\begin{enumerate}
\item Can our approach (compared to the others) achieve higher accuracy for equal computation time, or equal accuracy for smaller computation time? (Section~\ref{ssec:accuracy-speed})
\item How sensitive is the simulator accuracy to contact stiffness and damping? (Section~\ref{ssec:stiffness-damping})
\item What is the maximum integration time step that results in a \emph{stable}\footnote{We say that a simulation is ``stable'' if the robot state remains bounded.} motion? (Section~\ref{ssec:stability})
\item How accurately can~\eqref{eq:contact_dyn} predict future contact forces when assuming constant $A$ and $b$? (Section~\ref{ssec:force_prediction})
\item How much computation time is spent in the different operations of our simulator? Is there room for improvement? (Section~\ref{ssec:computation-times})
\end{enumerate}

\subsection{Accuracy Metric}
Following an approach similar to~\cite{Erez2015}, we measure accuracy with a \emph{local integration error}.
We compute the ground truth trajectory $x_q(t)$ using the simulator under analysis with an extremely small time step $\dt=1/64$ ms.
Let us define \mbox{$\hat{x}_q(t; t-\dt_c, x_q(t-\dt_c))$} as the state at time $t$ obtained by numerical integration starting from the ground-truth state $x_q(t-\dt_c)$, where $\dt_c (\ge \dt)$ is the time step of the controller.
We define the \emph{local} integration error as the error accumulated over one control time step:
$$
e(t) \triangleq \| x_q(t) \ominus \hat{x}_q(t; t-\dt_c, x_q(t-\dt_c) \|_{\infty}
$$
where $\ominus$ is a difference operator on the space of $x_q$.
In the numerical integration literature~\cite{Ascher1998} the \emph{local} integration (or truncation) error is typically defined using the integration step $\dt$ rather than the controller step $\dt_c$.
We chose to use the controller step to make errors comparable across tests with different integration steps (as in~\cite{Erez2015}).

\subsection{Test Description} 
\begin{table}[!tbp] 
\caption{Controller time steps.}
\centering 
\begin{tabular}{p{1.7cm} | p{1.0cm} || p{1.7cm} | p{1.0cm} }
\hline 
	Test		& $\dt_c$ [ms] & Test		& $\dt_c$ [ms] \\ \rowcolor[gray]{.9}
\hline 
	Solo-squat		& \changed{40}\ 		& Solo-trot			& 2 \\
	Solo-jump			& 10 		& Romeo-walk		& 40	 \\
[0.5ex] \hline 
\end{tabular} 
\label{tab:simu_params} 
\end{table}
To evaluate the trade-off between accuracy and computation time we tested each simulator with different time steps.
\changed{For \emph{Expo}, \emph{RK-4}, \emph{Eul-imp} we have started from $\dt = 1/8$ ms up to the controller time step $\dt = \dt_c$ with a logarithmic step of 2 (i.e. 1/8, 1/4, ..., $\dt_c/2$, $\dt_c$).
For \emph{Eul-exp} we have used the same approach, but starting from a value of $\dt$ resulting in roughly the same computation time of \emph{Expo}.}\~
For every test we have set $\dt_c$ to the largest value that still ensured control stability (see Table~\ref{tab:simu_params}).
Since our main interest lies in legged robots, our tests focused on quadrupeds and bipeds:
\begin{itemize}
\item \emph{Solo-squat}: Quadruped robot Solo~\cite{Grimminger2020} performing a squatting motion.
\item \emph{Solo-jump}: Quadruped robot Solo jumping in place. 
\item \emph{Solo-trot}: Quadruped robot Solo trotting forward.
\item \emph{Romeo-walk}: Humanoid robot Romeo~\cite{romeo} taking two walking steps.
\end{itemize}
It is important to note that the quadruped Solo has a total of 13 links, 18 degrees of freedom, 12 of which are actuated revolute joints, and 4 contact points (one on each foot). As for the humanoid Romeo, it consists of 32 links, 37 degrees of freedom, 31 actuated revolute joints and a total of 8 contact points (four on each foot).
In all tests, the control torques have been computed with a feedback controller, either a linear controller or a Task-Space Inverse Dynamics controller.
If not specified otherwise, we have used a contact stiffness $K=10^5$ N/m, and a contact damping $B=300$ Ns/m, which are reasonable values for contacts with a hard floor.
For homogeneity we have used the same value of friction coefficient $\mu=1$ across all our tests, even though this large friction was only needed for control stability of the quadruped jumping motion.
Besides testing the default \emph{Expo} simulator, we also tested 5 other versions of the same scheme where we used a reduced polynomial order in the Pad\'e approximant of the matrix exponential.
This leads to a reduced number of matrix-matrix multiplications (mmm), between 0 and 4 (see Section~\ref{ssec:boost_expm}).
This results in a faster but potentially less accurate computation of the matrix exponential.

All the code has been implemented in C++ and binded with Python.
For all dynamics computation we have used the Pinocchio library~\cite{Carpentier2019}.

\subsection{Accuracy-Speed Results}
\label{ssec:accuracy-speed}
\begin{figure*}[!tbp]
   \centering
   \subfloat[Solo squat]{\includegraphics[width=0.45\textwidth]{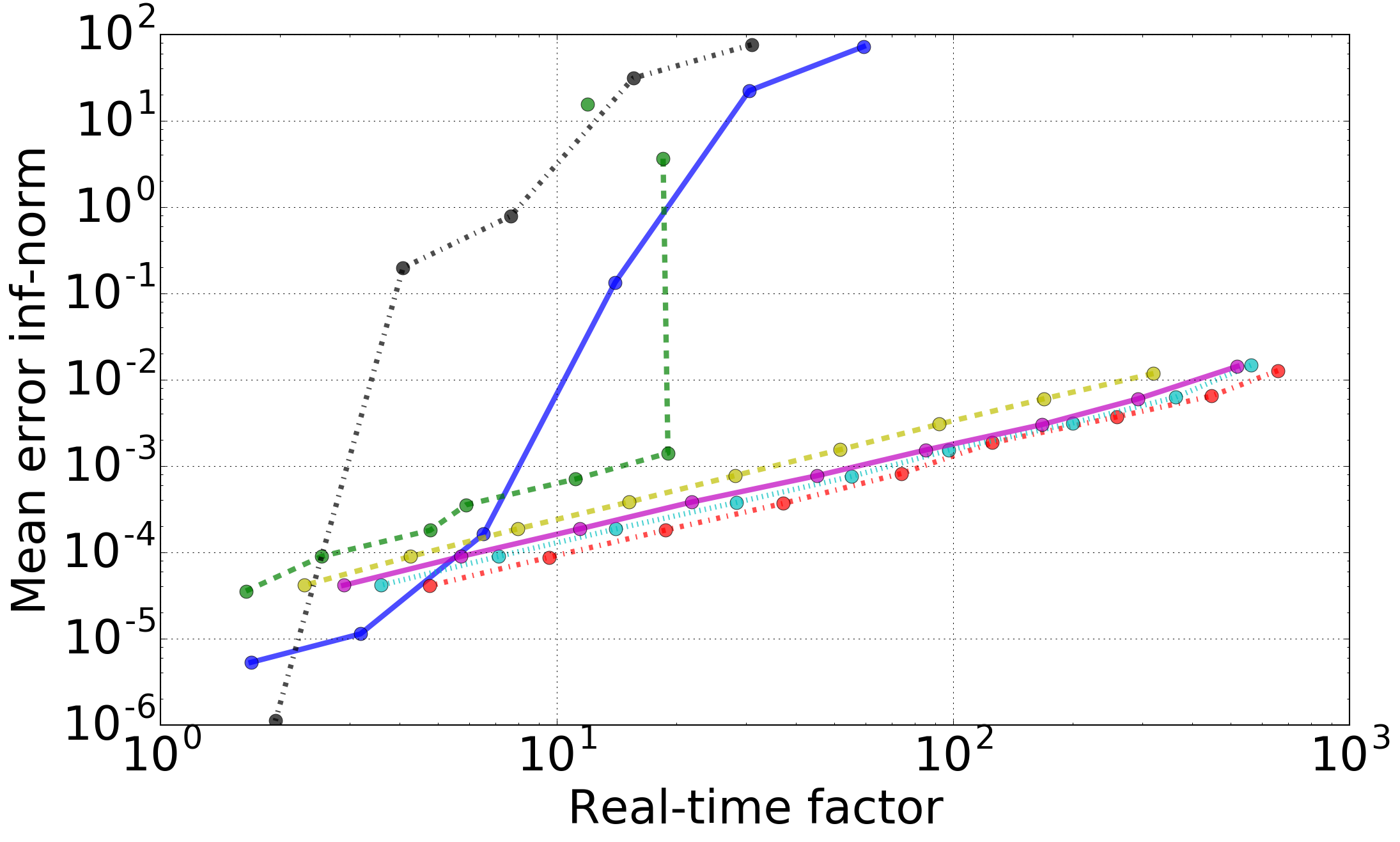}} \quad
   \subfloat[Solo jump]{\includegraphics[width=0.45\textwidth]{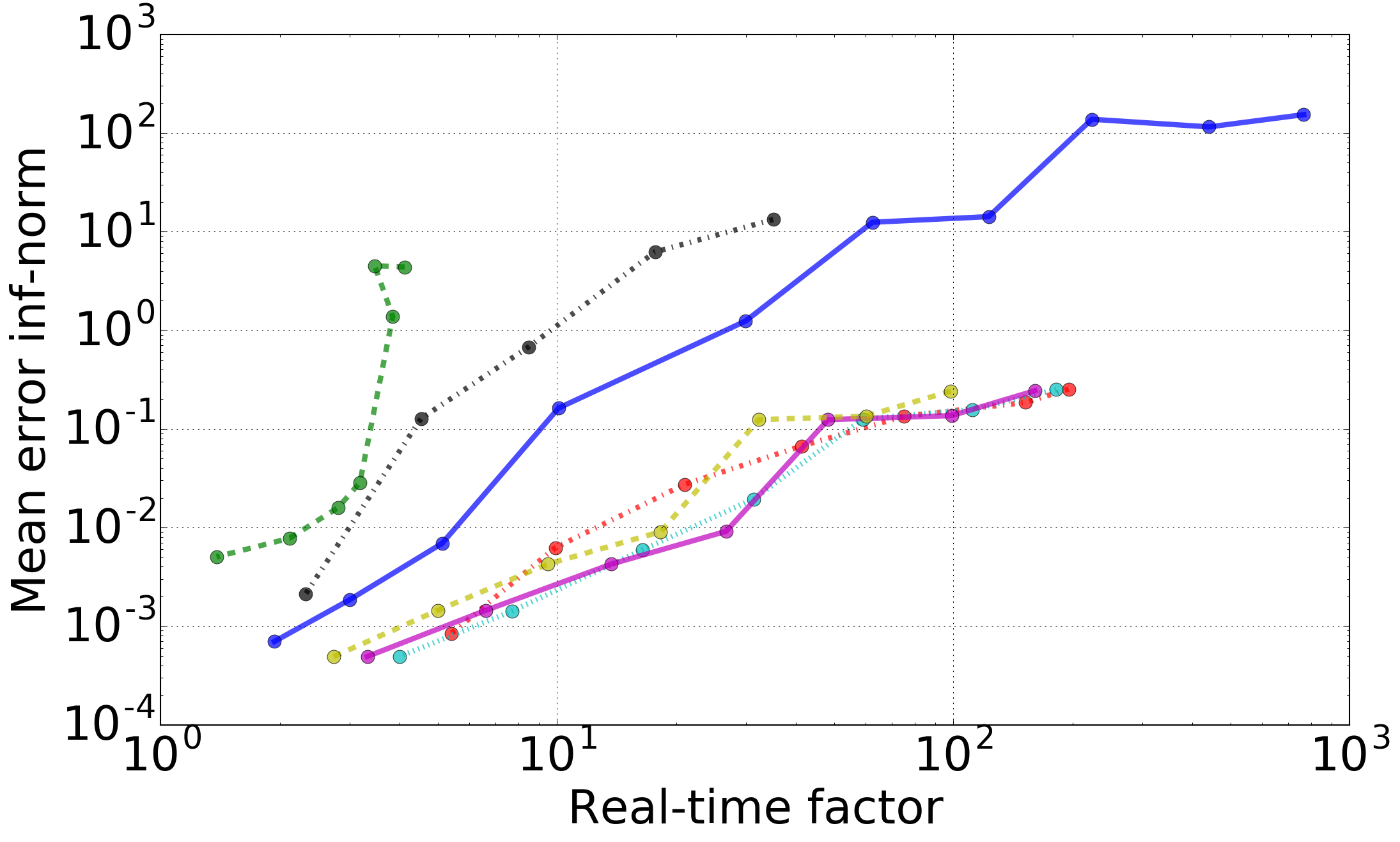}} \\
   \subfloat[Solo trot]{\includegraphics[width=0.45\textwidth]{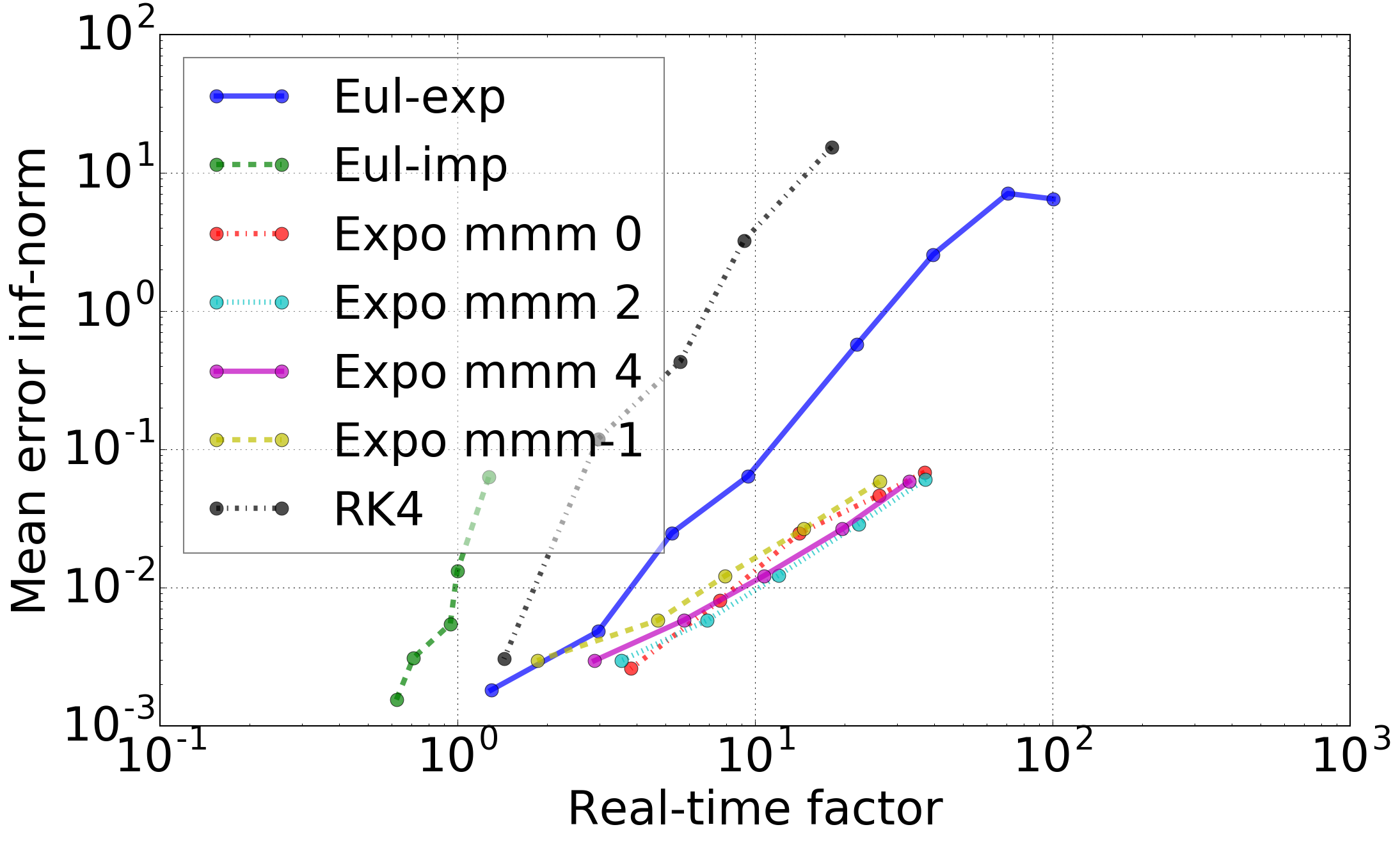}} \quad
   \subfloat[Romeo walk]{\includegraphics[width=0.45\textwidth]{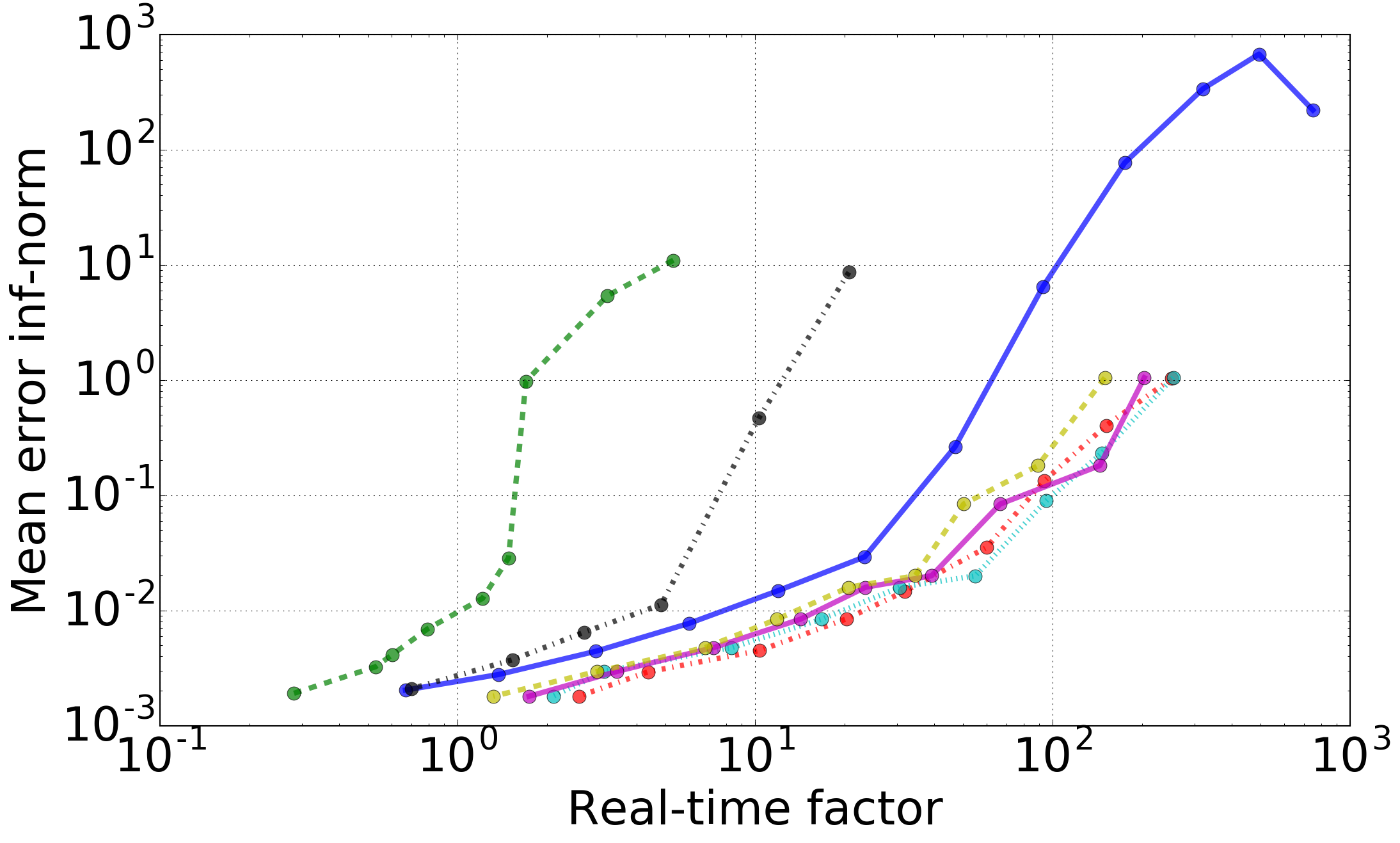}}
   \caption{Local integration errors vs real-time factors. The label \emph{mmm-1} in the legend corresponds to using the default number of matrix-matrix multiplications (mmm) in the computation of the matrix exponential.}
   \label{fig:local_error_vs_rt_factor}
\end{figure*}

\begin{figure*}[!tbp]
   \centering
   \subfloat[Solo squat]{\includegraphics[width=0.45\textwidth]{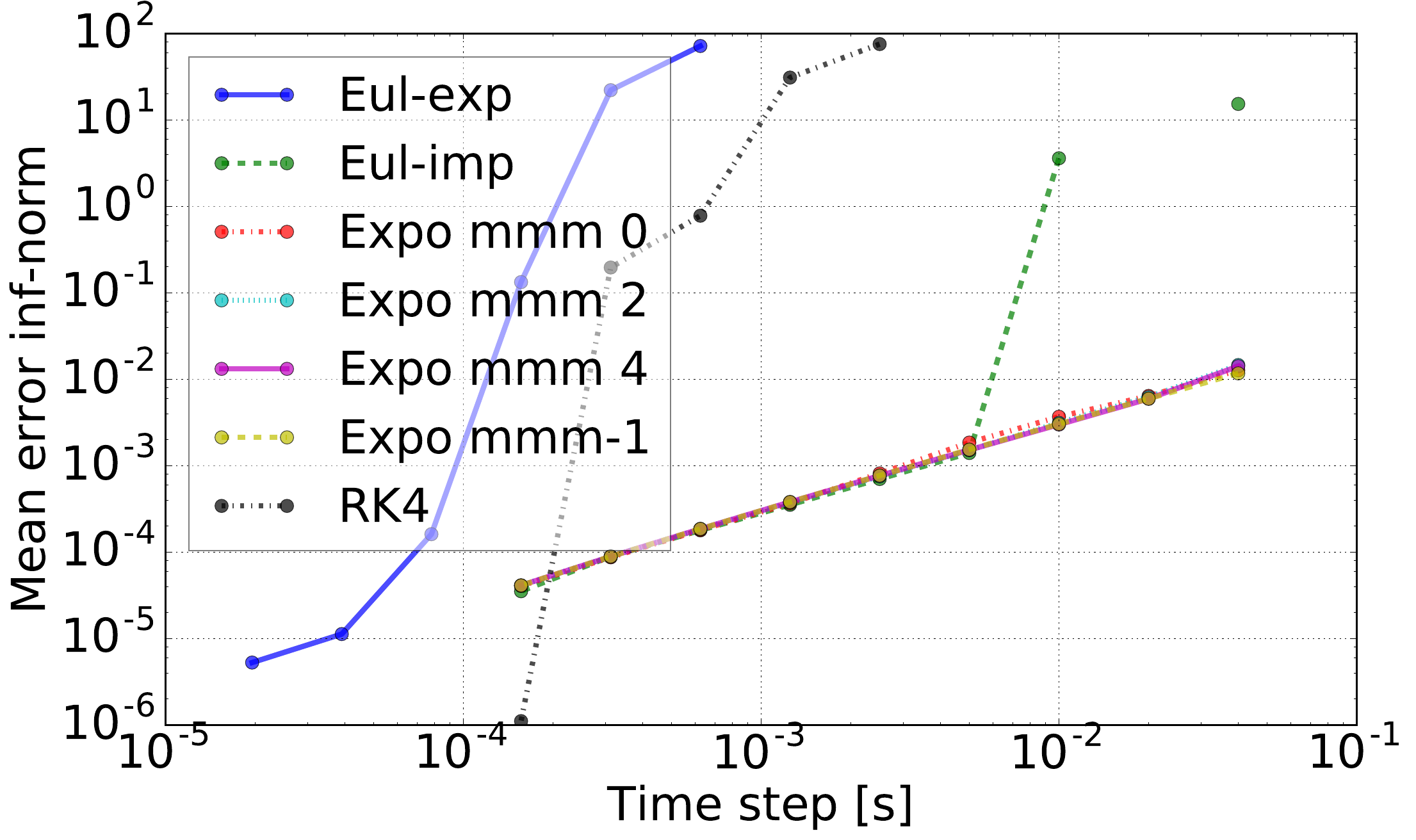}} \quad
   \subfloat[Solo jump]{\includegraphics[width=0.45\textwidth]{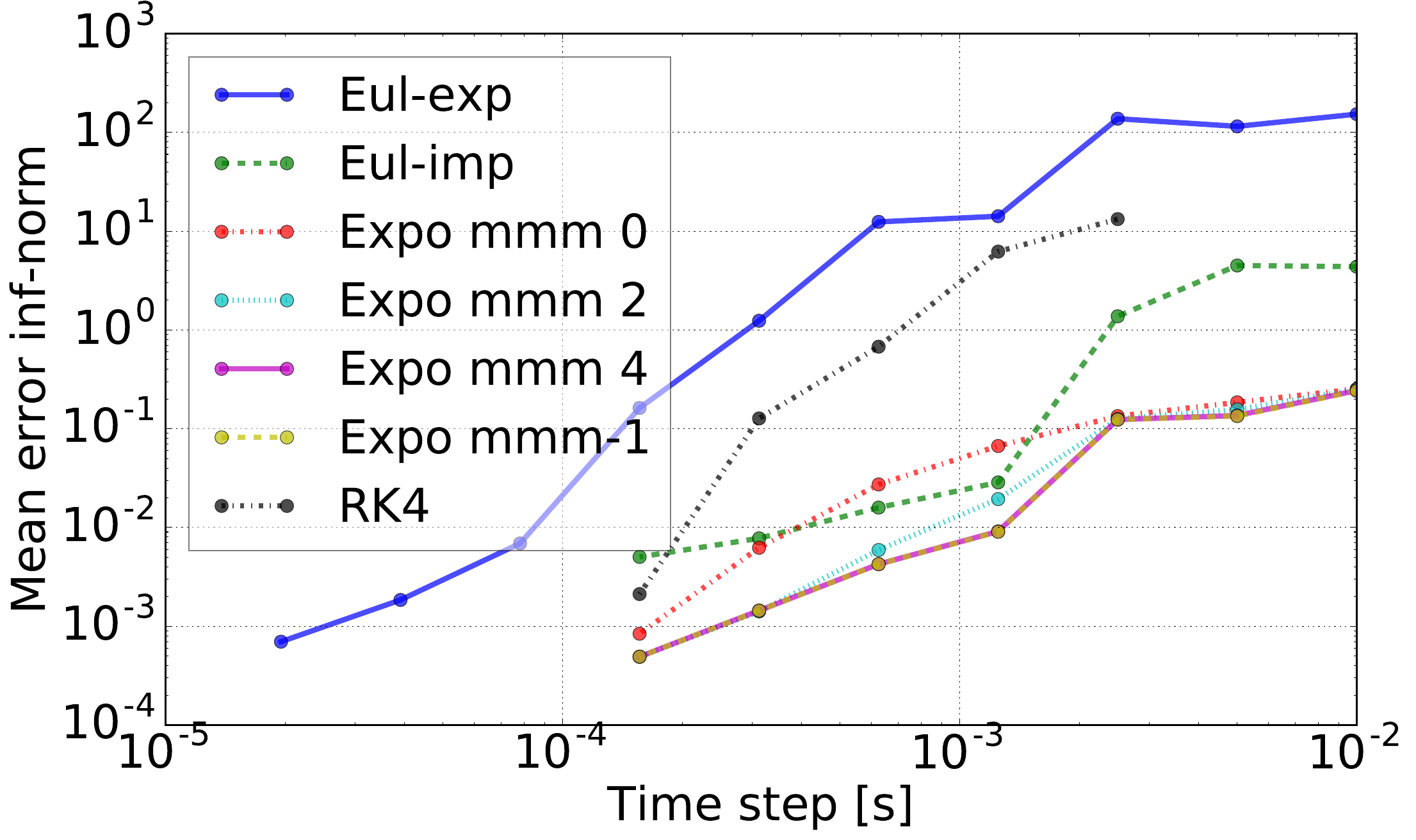}} \\
   \subfloat[Solo trot]{\includegraphics[width=0.45\textwidth]{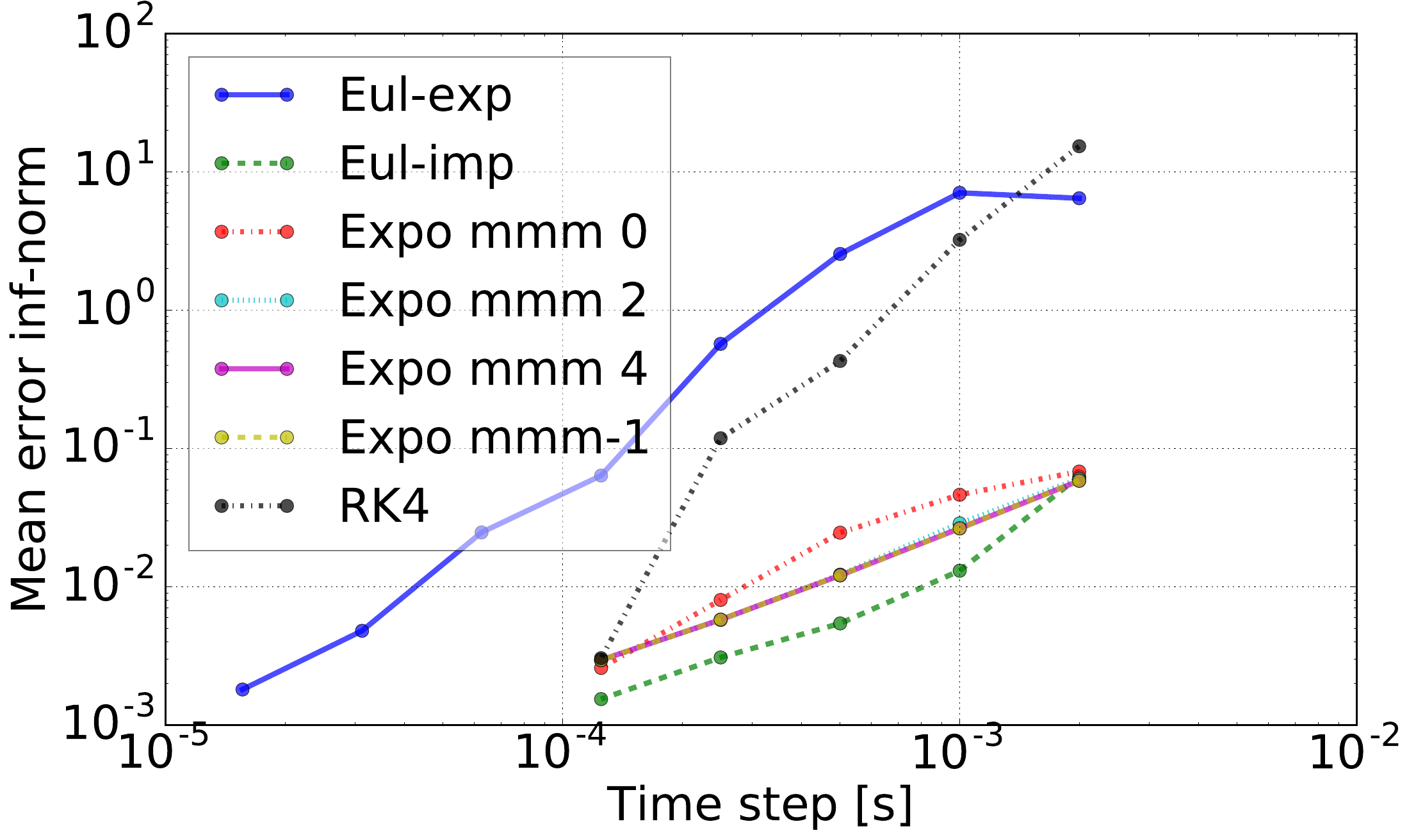}} \quad
   \subfloat[Romeo walk]{\includegraphics[width=0.45\textwidth]{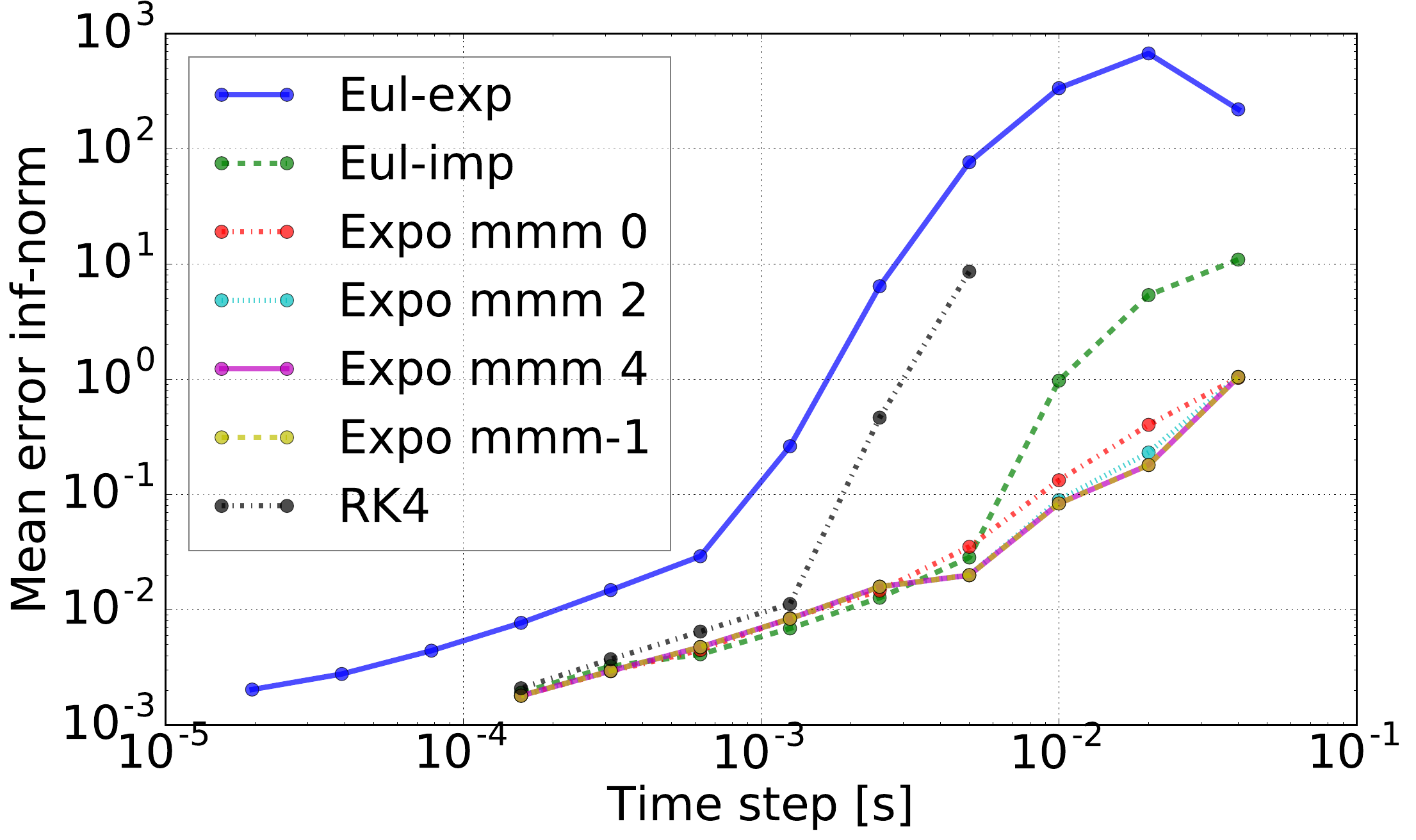}}
   \caption{Local integration errors vs integration time step.}
   \label{fig:local_error_vs_dt}
\end{figure*}

Fig.~\ref{fig:local_error_vs_rt_factor} and \ref{fig:local_error_vs_dt} summarize the results for the four tests.
Fig.~\ref{fig:local_error_vs_rt_factor} plots \emph{local errors} vs \emph{real-time factor}, which measures how many times the simulation was faster than real time.
Fig.~\ref{fig:local_error_vs_dt} instead plots \emph{local errors} vs \emph{integration time step}.
Even though our main interest is in the trade-off between computation time and accuracy, which is depicted in Fig.~\ref{fig:local_error_vs_rt_factor}, we decided to report also the accuracy as a function of integration time in Fig.~\ref{fig:local_error_vs_dt}, to provide more information about the behavior of the different methods.

Overall, \emph{Expo} outperformed the other methods in all tests, showing faster computation for equal accuracy, or greater accuracy for equal computation time.
Surprisingly, the second best method overall was the simple \emph{Eul-exp}, even though it was partially beaten by \emph{Eul-imp} in ``solo-squat''.
\emph{RK4} was comparable to \emph{Eul-exp} for small time steps, but surprisingly worse for large time steps.
These results show a sudden increase of integration error of \emph{Eul-exp} and \emph{RK4} for large real-time factors---corresponding to large $\dt$.
This is because of the poor stability of explicit methods.

\emph{Eul-imp} sometimes failed to converge to the desired error threshold ($10^{-6}$), in particular when using large integration steps.
This is not surprising because the system dynamics is non-continuous at impacts, and \emph{Eul-imp} uses a gradient-based method (Newton) that is suited for smooth systems.
Despite this, Fig.~\ref{fig:local_error_vs_dt} shows that in most cases \emph{Eul-imp} gives integration errors similar to \emph{Expo} for the same integration time step.
More precisely, \emph{Eul-imp} is almost indistinguishable from \emph{Expo} in ``solo-squat'' for $\dt \le 5$ ms, whereas it is unstable for larger $\dt$.
\emph{Eul-imp} is a bit worse than \emph{Expo} in ``solo-jump'' and ``romeo-walk'' (but only for large time steps), and slightly better in ``solo-trot''.
Overall, \emph{Eul-imp} performed similarly to \emph{Expo} for the same $\dt$, but resulted in much larger computation times, making it often the worst one in terms of accuracy-speed trade-off.

\emph{Expo} instead shows a graceful degradation of accuracy for large real-time factors, making it an excellent candidate for fast low-accuracy simulations, which are typically desirable in MPC.
In general, the \emph{Expo} versions using a reduced number of mmm outperformed the standard \emph{Expo}, but which number of mmm is optimal depends on the specific test and time step.
As expected, when $\dt$ is smaller we can use a lower number of mmm.
Automatically finding the optimal number of mmm is an interesting direction for future work.

\subsection{Stiffness and Damping}
\label{ssec:stiffness-damping}
\begin{figure}[!tbp]
   \centering
   \subfloat[Varying contact stiffness (with fixed damping ratio of 0.5)]{\includegraphics[width=0.4\textwidth]{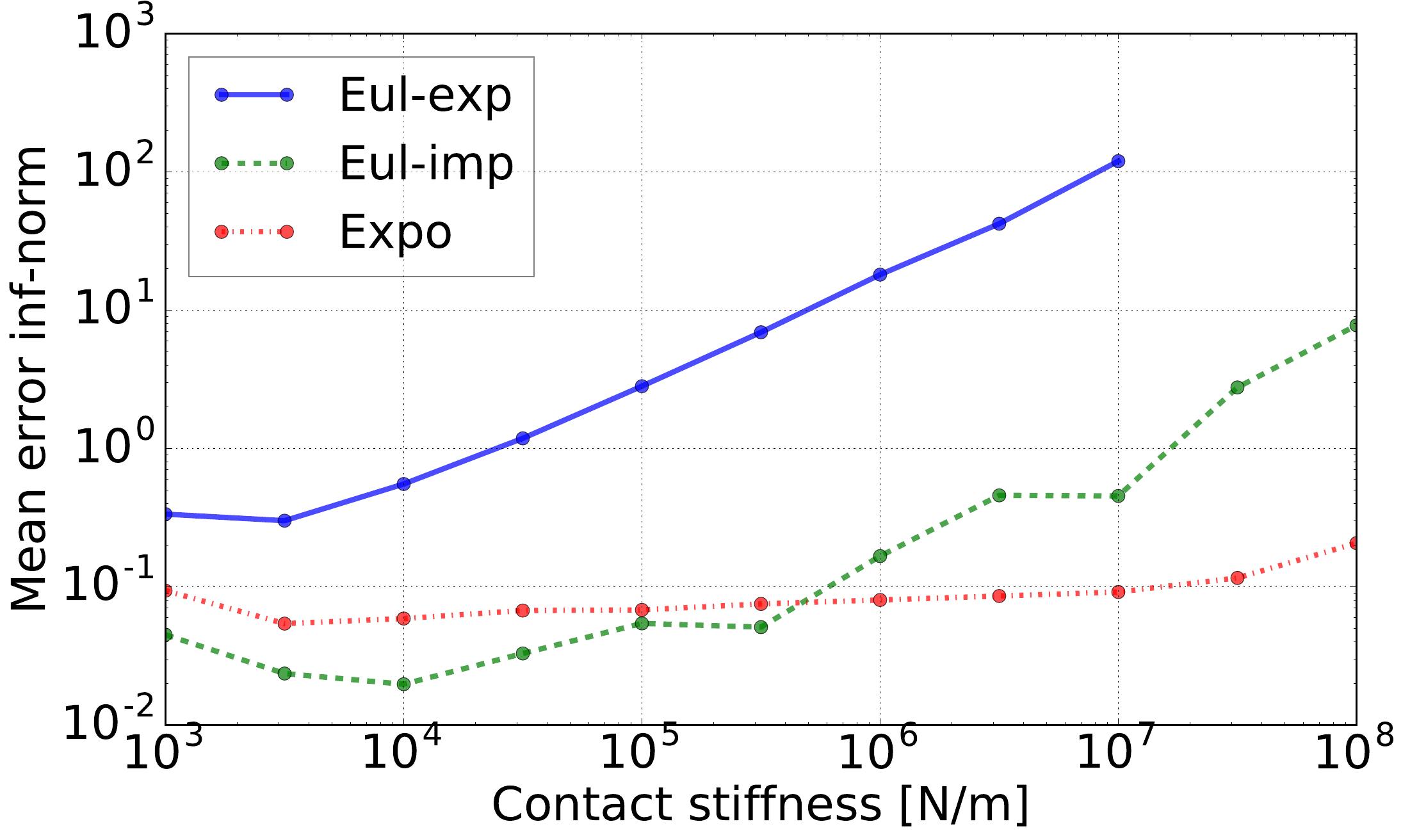}} \quad
   \subfloat[Varying contact damping ratio (with fixed stiffness of $10^5$ N/m).]{\includegraphics[width=0.4\textwidth]{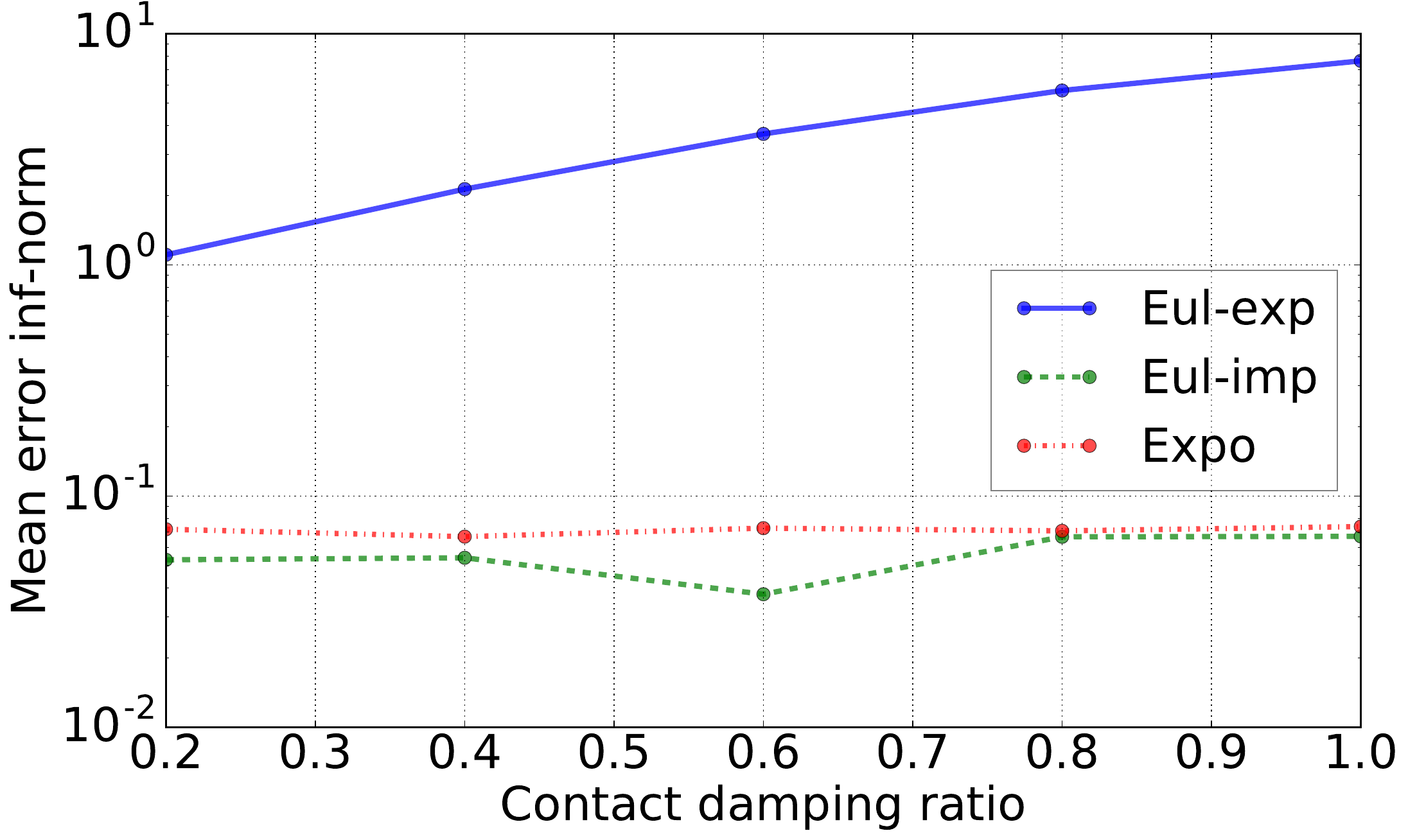}}
   \caption{Local integration errors vs contact stiffness and damping ratio for the ``solo trot'' test using a fixed \changed{integration time step for \emph{Eul-exp} (1/2 ms), \emph{Expo} (2 ms) and \emph{Eul-imp} (2 ms).}}
   \label{fig:local_errorvs_stiff_damp}
\end{figure}
This subsection investigates the sensitivity to contact stiffness and damping ratio of \emph{Expo}, \emph{Eul-exp} \changed{and \emph{Eul-imp}.}\~
The damping ratio is defined as $\frac{B}{2 \sqrt{K}}$.
A damping ratio of 1 corresponds to a \emph{critically damped} contact.
These results are based on the ``solo-trot'' scenario.
\changed{For \emph{Eul-exp} we have used $\dt$=1/2 ms (real-time factor $\approx$50). 
Then, we have set $\dt$=2 ms for \emph{Expo} 
so that it had roughly the same computation time, and $\dt$=2 ms for \emph{Eul-imp}, so that it performed similarly to \emph{Expo} (even though with much larger computation times).}\~

Fig.~\ref{fig:local_errorvs_stiff_damp} shows the local integration error as we vary the contact stiffness (with fixed damping ratio) and the damping ratio (with fixed contact stiffness).
\changed{\emph{Expo} performs consistently as damping ratio and stiffness increase up to \mbox{$K=10^8$}, which roughly corresponds to a ground penetration of 0.01 mm for 100 kg of weight on a single contact point. 
The error of \emph{Eul-exp} instead is highly affected by both stiffness and damping.
\emph{Eul-imp} performed slightly better than \emph{Expo} in most cases (but at the cost of being 50 times slower), except for very stiff contacts ($K\ge10^6$), where it led to larger errors.}

\subsection{Stability}
\label{ssec:stability}
\begin{table}[!tbp] 
\caption{Maximum integration time steps to achieve a stable motion.}
\centering 
\begin{tabular}{p{1.9cm} | p{1.6cm} | p{2cm} | p{2cm}}
\hline 
	Test		& \emph{Expo} $\dt$ [ms] & \emph{Eul-exp} $\dt$ [ms] & \changed{\emph{Eul-imp}}\~ $\dt$ [ms] \\ \rowcolor[gray]{.9}
\hline 
	Solo-squat		& 40 			& 40/512 $\approx$ 0.08 & 40/2 = 20 \\
	Solo-jump			& 10 			& 10/64 $\approx$ 0.16 & {10/2 = 5}	 \\ \rowcolor[gray]{.9}
	Solo-trot			& 2			& 2/16  $\approx$ 0.13 & {2} \\  
	Romeo-walk		& 40/8 = 5		& 40/32 = 1.25 & {40/4 = 10}	\\
[0.5ex] \hline 
\end{tabular} 
\label{tab:stability} 
\end{table}
To test the stability of the simulators we have repeated the previous tests, but without resetting the state to the ground truth after every control loop.
Table~\ref{tab:stability} reports, \changed{for \emph{Expo}, \emph{Eul-exp} and \emph{Eul-imp}, the largest integration time step for which the system remained stable.
\emph{Expo} and \emph{Eul-imp} showed similar stability, both remaining stable for large time steps, mostly between 5 and 40 ms.
In the tests ``solo-squat'' and ``solo-jump'' \emph{Expo} was stable even with a larger time step than \emph{Eul-imp}, whereas the opposite happened in ``romeo-walk''.
\emph{Eul-exp} instead showed poor stability, needing a time step between 4 and 512 times smaller than \emph{Expo} to remain stable.}

\subsection{Force Prediction}
\label{ssec:force_prediction}
\begin{figure}[!tbp]
   \centering
   \subfloat[Contact velocity at impact: 0.1 m/s.]{\includegraphics[width=0.45\textwidth]{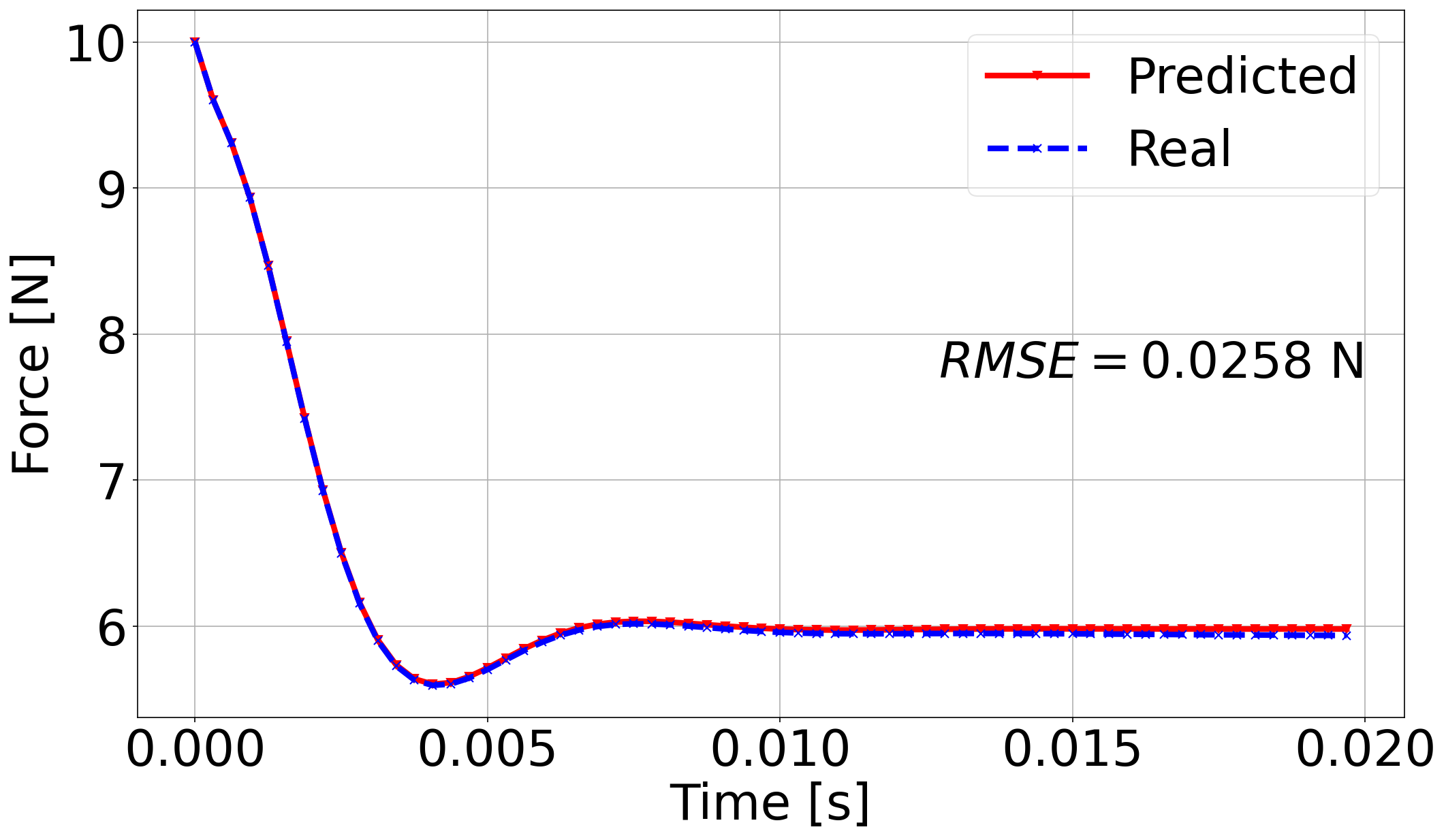}} \quad
   \subfloat[Zero contact velocity at impact.]{\includegraphics[width=0.45\textwidth]{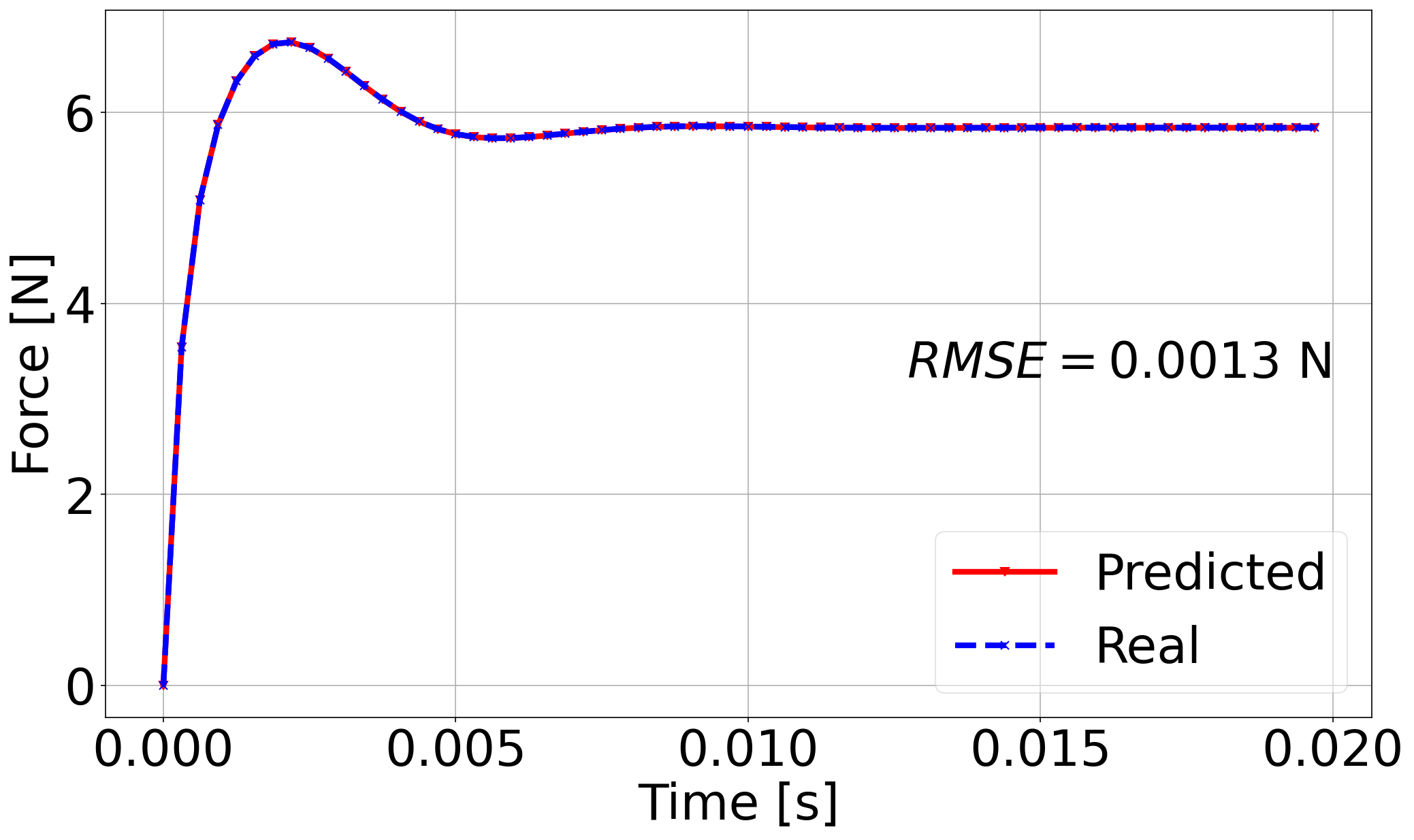}}
   \caption{Comparison of contact forces in normal direction with forces predicted using matrix exponential.}
   \label{fig:force_prediction}
\end{figure}
\begin{figure}[!tbp]
   \centering
   \subfloat[Quadruped Trotting.]{\includegraphics[width=0.45\textwidth]{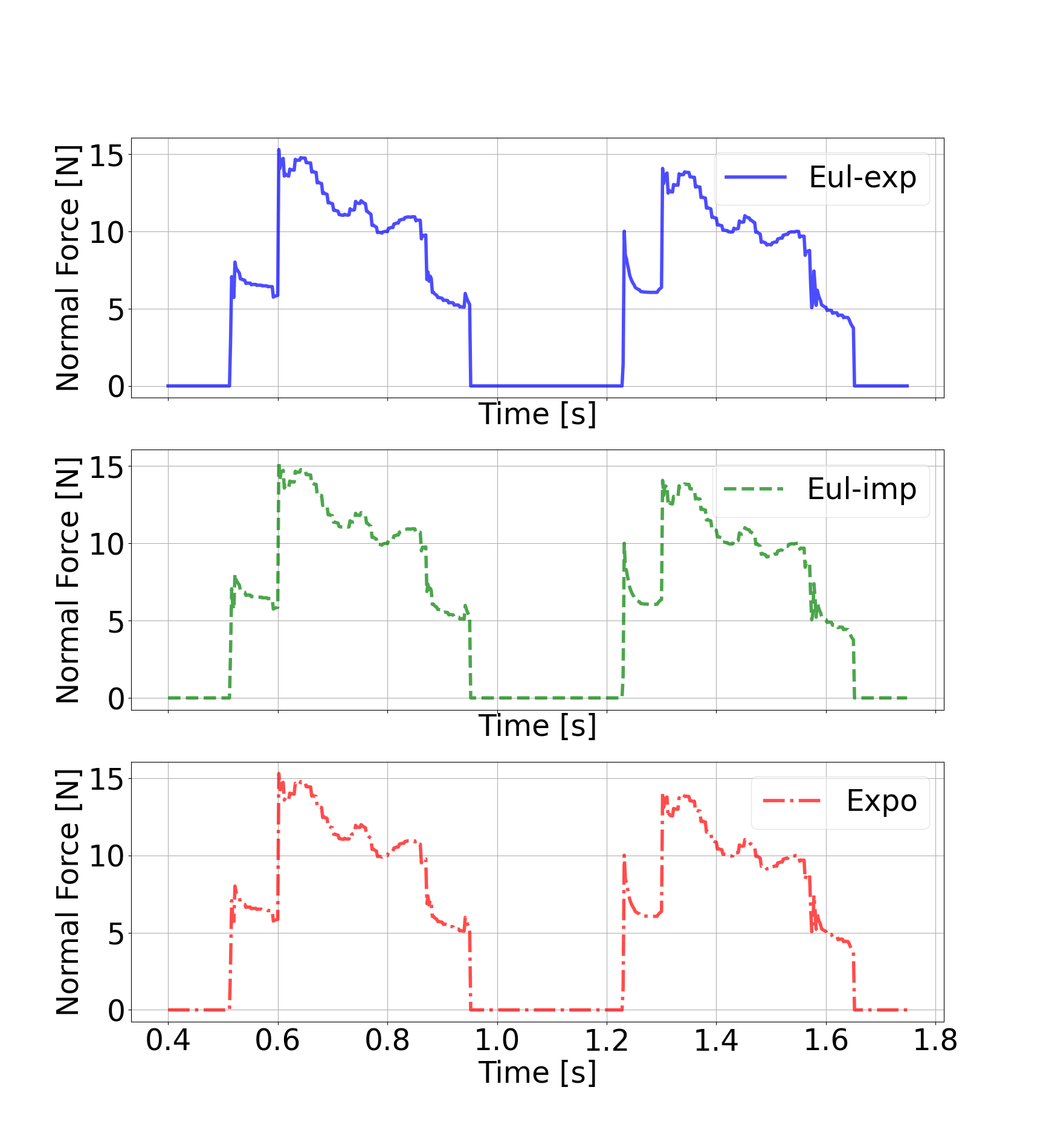}} \quad
   \subfloat[Biped Walking.]{\includegraphics[width=0.45\textwidth]{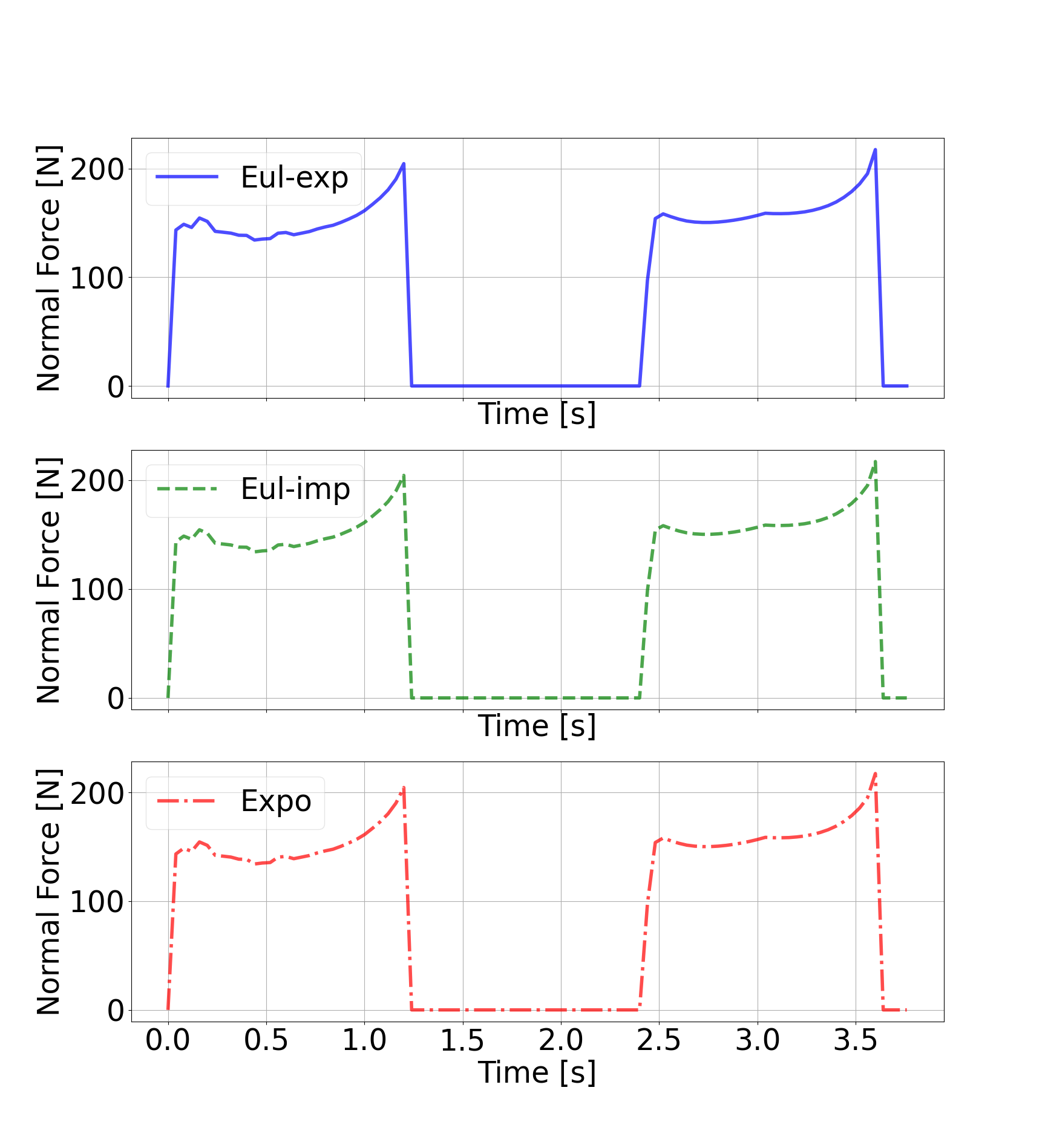}}
   \caption{Normal contact force during quadruped trotting and biped walking.}
   \label{fig:walking_force}
\end{figure}

To gain some insights into the internal computations of \emph{Expo}, we show in Fig.~\ref{fig:force_prediction} the normal contact forces predicted with~\eqref{eq:contact_dyn} assuming constant $A$ and $b$---which is the key assumption of our method.
Since $A$ and $b$ depend on $q$ and $v$, which vary during the time step, one could expect that neglecting their variations would result in significant force prediction errors.
However, Fig.~\ref{fig:force_prediction} shows that the force prediction can be accurate over a rather long time horizon (20 ms).
These forces were generated at the beginning of the ``solo-squat'' test, using different initial velocities. 
Since a linear spring-damper model is used, a sudden discontinuity in the contact force is expected when a point reaches contact with a non-zero velocity. 
To demonstrate this, the normal contact force at a single contact point is plotted for the cases of the trotting quadruped and the walking biped in Fig.~\ref{fig:walking_force}. 
Depending on the velocity at contact time, a finite jump in the contact force is observed.

\subsection{Computation Times}
\label{ssec:computation-times}
\begin{table}[!tbp] 
\caption{Computation times of \emph{Expo} for ``solo-trot'', using zero mmm for the matrix exponential (in parentheses the values using the standard matrix exponential routine).}
\centering 
\begin{tabular}{p{3.0cm} | p{1.5cm} | p{2.0cm} }
\hline 
	Operation  		& Mean Time [$\mu$s] & Percentage of Total Time [\%]\\ \rowcolor[gray]{.9}
\hline 
                          step 	&   39  (94) &  100 \\
  computeIntegrals 		&   13 (67)  &   33 (72) \\ \rowcolor[gray]{.9}
       prepareExpLDS 		&   13   &   33 (14)\\
computeContactForces 	&    8    &   20 (8) \\ \rowcolor[gray]{.9}
\emph{Eul-exp} step 			&    9 	& - \\
[0.5ex] \hline 
\end{tabular} 
\label{tab:computation_times} 
\end{table}
We report here a breakdown of the computation time of our method.
The times shown in Tab.~\ref{tab:computation_times} are for the ``solo-trot'' test, which means that $v \in \Rv{18}$ and most of the times $\lambda \in \Rv{6}$.
Most computation time (86\%) is spent in three operations: \texttt{computeIntegrals}, \texttt{prepareExpLDS}, and \texttt{computeContactForces}.
\texttt{computeIntegrals} boils down to computing a matrix exponential.
This takes 72\% of the total time when using a standard \texttt{expm} routine (without balancing and reduced matrix-matrix multiplications), but it goes down to 33\% with our optimized version using zero matrix-matrix multiplications---we have seen in Fig.~\ref{fig:local_error_vs_rt_factor} that often this results in only a small loss of integration accuracy.
The preparation of the linear dynamical system~\eqref{eq:contact_dyn} (\texttt{prepareExpLDS}, which includes the computations of $h(q,v)$ with RNEA, $M(q)$ with CRBA, and $\Upsilon$ with a custom sparse Cholesky decomposition) takes an equal amount of time: 13 $\mu$s on average, namely 33\% of the total time.
The third operation (\texttt{computeContactForces}) takes 20\% of the total time, and it includes the computation of all kinematic quantities (contact point positions, velocities, accelerations, Jacobian) and the contact detection.

We believe that computation times could be improved, especially for the first two operations.
In \texttt{computeIntegrals} we could test novel techniques~\cite{Sastre2019} to compute the matrix exponential, exploit the sparse structure of the matrix $A$, and warm-start the computation using quantities computed at the previous cycle.
In \texttt{prepareExpLDS}, the inverse contact-space inertia matrix $\Upsilon$ could be computed faster using a customized algorithm, rather than with products between $J$, $M^{-1}$ and $J^\top$~\cite{Featherstone2009}.
Overall, it seems impossible to reach the same efficiency of a simple \emph{Eul-exp} step (9 $\mu$s), but we think we could reach computation times in the range [20, 30] $\mu$s.

%% file: sections/conclusions.tex
\section{Conclusions}
\label{sec:conclusions}
This paper has presented a new approach to simulate articulated systems subject to stiff visco-elastic frictional contacts.
The novelty of the approach lies in the numerical integration, which applies a first-order Exponential Integrator scheme to the contact point dynamics to obtain a time-varying expression of the contact forces.
These contact forces are then integrated analytically, exploiting theoretical results on the integrals of the matrix exponential~\cite{Carbonell2008}, and advanced numerical algorithms for its fast computation~\cite{Higham2005}.
Comparison with standard integration schemes, both implicit and explicit, highlighted the benefits of the proposed approach in terms of speed-accuracy trade off, and stability.
\changed{Overall, the proposed approach performed similarly to an implicit scheme in terms of stability and accuracy, but without the excessive computational burden.}\~

Given its good behavior in the high-speed/low-accuracy regime, we believe that this simulation technique could be an excellent candidate for MPC.
To do that, we will need to differentiate the integration scheme, which should be feasible.
We also plan to investigate the improvement, in terms of computational efficiency, of the needed dynamics quantities and the matrix exponential.